\documentclass{article}

\usepackage{microtype}
\usepackage{graphicx}
\usepackage{subcaption}
\usepackage{booktabs} 

\usepackage{hyperref}

\usepackage[preprint]{icml2026}

\usepackage{hyperref}
\usepackage{url}

\usepackage{microtype}
\usepackage{hyperref}
\usepackage{url}
\usepackage{booktabs}
\usepackage{times}
\usepackage{latexsym}
\usepackage{amsmath}
\usepackage{amsfonts}
\usepackage{graphicx}
\usepackage{multirow}
\usepackage{nicematrix}
\usepackage{pgfplots}
\pgfplotsset{compat=1.17}
\usepackage{subcaption}
\usepackage[T1]{fontenc}

\usepackage{tikz}

\usepackage{amsmath}
\usepackage{amssymb}
\usepackage{mathtools}
\usepackage{amsthm}

\usepackage{listings}
\usepackage{color}
\usepackage{minted}
\definecolor{dkgreen}{rgb}{0,0.6,0}
\definecolor{gray}{rgb}{0.5,0.5,0.5}
\definecolor{mauve}{rgb}{0.58,0,0.82}
\definecolor{lightblue}{rgb}{0.9, 0.95, 1.0}
\definecolor{lightgray}{rgb}{0.9, 0.9, 0.9}

\definecolor{-}{rgb}{0.70,0.13,0.13}
\definecolor{+}{rgb}{0.0, 0.6, 0.3} 

\definecolor{thinkcolor}{RGB}{0,139,139}      
\definecolor{searchcolor}{RGB}{255,140,0}     
\definecolor{infocolor}{RGB}{128,0,128}       
\definecolor{answercolor}{RGB}{0,128,128}     
\definecolor{questioncolor}{RGB}{220,20,60}   

\usepackage{enumitem}

\usepackage{colortbl}
\usepackage{pifont}
\usepackage{algorithmic}
\usepackage{algorithm}

\usepackage[utf8]{inputenc}
\usepackage{multirow}
\usepackage{microtype}
\usepackage{booktabs}

\usepackage{wrapfig}    

\usepackage{hyperref}
\usepackage{url}

\usepackage{pifont}
\usepackage{microtype}
\usepackage{hyperref}
\usepackage{url}
\usepackage{booktabs}
\usepackage{times}
\usepackage{latexsym}
\usepackage{amsmath}
\usepackage{amsfonts}
\usepackage{graphicx}
\usepackage{multirow}
\usepackage{microtype}
\usepackage{booktabs}
\usepackage{graphicx}
\usepackage{subcaption}
\usepackage[dvipsnames]{xcolor}
\usepackage{wrapfig}

\usepackage[capitalize,noabbrev]{cleveref}

\theoremstyle{plain}

\theoremstyle{definition}

\theoremstyle{remark}

\usepackage[textsize=tiny]{todonotes}

\icmltitlerunning{In-Context Reinforcement Learning for Tool Use in Large Language Models}

\begin{document}

\twocolumn[
  \icmltitle{In-Context Reinforcement Learning for Tool Use in Large Language Models}



  \icmlsetsymbol{equal}{*}

  \begin{icmlauthorlist}
    \icmlauthor{Yaoqi Ye}{equal,nus}
    \icmlauthor{Yiran Zhao}{equal,sf}
    \icmlauthor{Keyu Duan}{nus}
    \icmlauthor{Zeyu Zheng}{uc1}
    \icmlauthor{Kenji Kawaguchi}{nus}
    \icmlauthor{Cihang Xie}{uc2}
    \icmlauthor{Michael Qizhe Shieh}{nus}
  \end{icmlauthorlist}

  \icmlaffiliation{nus}{National University of Singapore}
  \icmlaffiliation{sf}{Salesforce AI Research}
  \icmlaffiliation{uc1}{University of California Berkeley}
\icmlaffiliation{uc2}{University of California, Santa Cruz}

  \icmlcorrespondingauthor{Yiran Zhao}{zhaoyiran0924@gmail.com}
  \icmlcorrespondingauthor{Michael Qizhe Shieh}{michaelshieh@comp.nus.edu.sg}

  \icmlkeywords{Machine Learning, ICML}

  \vskip 0.3in
]



\printAffiliationsAndNotice{\icmlEqualContribution}

\begin{abstract}


While large language models (LLMs) exhibit strong reasoning abilities, their performance on complex tasks is often constrained by the limitations of their internal knowledge. A compelling approach to overcome this challenge is to augment these models with external tools—such as Python interpreters for mathematical computations or search engines for retrieving factual information. However, enabling models to use these tools effectively remains a significant challenge. Existing methods typically rely on cold-start pipelines that begin with supervised fine-tuning (SFT), followed by reinforcement learning (RL). These approaches often require substantial amounts of labeled data for SFT, which is expensive to annotate or synthesize. In this work, we propose \underline{I}n-\underline{C}ontext \underline{R}einforcement \underline{L}earning (\texttt{ICRL}), an RL-only framework that eliminates the need for SFT by leveraging few-shot prompting during the rollout stage of RL. Specifically, \texttt{ICRL} introduces in-context examples within the rollout prompts to teach the model how to invoke external tools. Furthermore, as training progresses, the number of in-context examples is gradually reduced, eventually reaching a zero-shot setting where the model learns to call tools independently. We conduct extensive experiments across a range of reasoning and tool-use benchmarks. Results show that \texttt{ICRL} achieves state-of-the-art performance, demonstrating its effectiveness as a scalable, data-efficient alternative to traditional SFT-based pipelines. \footnote{Code is publicly available at \url{https://github.com/applese233/ICRL}}

\end{abstract}

\section{Introduction}

Recent advances in large language models (LLMs)~\citep{guo2025deepseek, yang2025qwen3, seed2025seed1, team2025kimi} have shown their effectiveness in addressing a wide range of complex tasks~\citep{wang2024qimprovingmultistepreasoning, hsiao2025a, shi2025toollearningwildempowering, Qu2025}. Nevertheless, a key limitation remains: these models rely on a fixed body of knowledge acquired during pretraining, which inherently restricts their ability to adapt to new or time-sensitive information~\citep{gao2023retrieval, zhu2025evolvebench, wang2024knowledge, cheng2024dated, matarazzo2025survey}. To mitigate this issue and enhance model flexibility, recent research has focused on enabling LLMs to interact with external tools during inference. This includes generating and executing Python code for mathematical reasoning, leveraging web search engines to access up-to-date and domain-specific content, and invoking dedicated helper models for specialized subtasks~\citep{guo2024large, team2025kimi, li2025webthinker, jin2025search, feng2025retool}.

The dominant training paradigms for LLMs either leverage reinforcement learning (RL) with verifiable reward signals~\citep{guo2025deepseek, jin2025search, zhao2025parallelsearch}, or adopt a cold-start strategy that begins with supervised fine-tuning (SFT) followed by an RL phase~\citep{mei20252, nguyen2025sfr}. Directly applying RL from scratch often yields poor performance, as the model lacks initial tool-use abilities and struggles with ineffective exploration. While incorporating an SFT stage can guide the model toward a more favorable initialization, it typically requires a large amount of high-quality labeled data, which is expensive to annotate or synthesize.

\begin{figure*}[t]
  \centering
\includegraphics[width=0.98\textwidth]{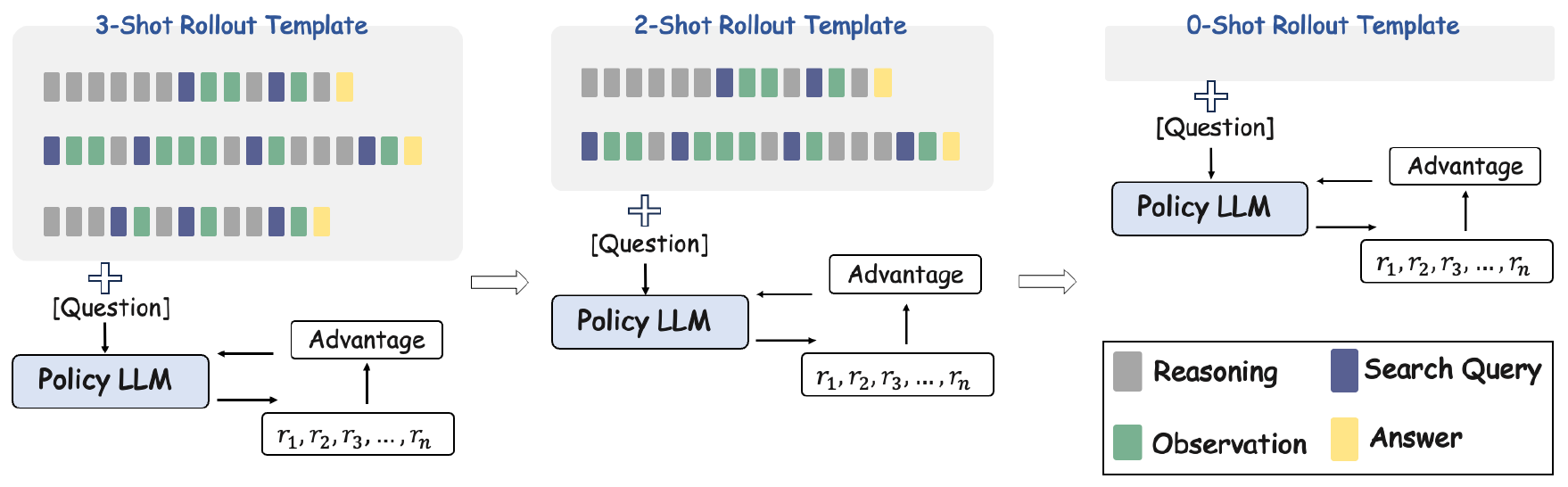}
  \caption{\texttt{ICRL} training workflow. The model is trained through a multi-stage curriculum that gradually reduces the number of in-context examples in the rollout template. At each stage, the LLM generates tool-augmented rollouts, receives rewards, and updates its policy via reinforcement learning, enabling a transition from imitation to autonomous tool use.}
  \label{fig:framework}
\end{figure*}

In this work, we introduce \underline{I}n-\underline{C}ontext \underline{R}einforcement \underline{L}earning (\texttt{ICRL}), a lightweight and supervision-efficient framework for training LLMs to perform tool-augmented reasoning. Unlike prior approaches that rely on SFT, \texttt{ICRL} teaches tool use directly through RL rollouts that are augmented with in-context demonstrations. Specifically, during RL training, we construct each rollout prompt by prepending a small number of few-shot examples that illustrate how to reason step-by-step, invoke tools in a structured format, and generate final answers. These demonstrations serve as soft supervision during exploration, guiding the model toward successful behavior without requiring labeled trajectories. Furthermore, as training progresses, we gradually reduce the number of demonstrations included in these rollout prompts, transitioning the model from few-shot to zero-shot settings. This progressive reduction forms a curriculum that encourages the model to internalize tool-use strategies and produce structured outputs autonomously, without relying on prompt-based scaffolding. We optimize the model using RL with a reward that balances task accuracy and format correctness. To ensure stability, we adopt GRPO~\citep{shao2024deepseekmath} with loss masking to ignore non-trainable tool outputs. By embedding and gradually removing demonstrations from the RL rollouts, \texttt{ICRL} merges the efficiency of prompting with the adaptability of RL, offering a scalable, supervision-light alternative to traditional SFT+RL pipelines.

We conduct comprehensive experiments across a range of QA and reasoning benchmarks to evaluate the effectiveness of \texttt{ICRL}. Without relying on supervised fine-tuning or ground-truth tool traces, \texttt{ICRL} achieves state-of-the-art performance on challenging QA datasets, outperforming strong baselines such as ZeroSearch~\citep{sun2025zerosearch}, Search-R1~\citep{jin2025search}, and ParallelSearch~\citep{zhao2025parallelsearch} by up to 8.9 on Qwen2.5-3B~\citep{qwen2.5} and 7.3 on Qwen2.5-7B in average exact match accuracy. The gains are especially pronounced on multi-hop reasoning tasks, where \texttt{ICRL} achieves double-digit improvements on datasets like TriviaQA~\citep{joshi2017triviaqa}, 2Wiki~\citep{ho2020constructing}, and Musique~\citep{trivedi2022musique}. Furthermore, in contrast to methods such as O$^2$-Searcher~\citep{mei20252} that require cold-start SFT to learn complex tool-use behavior, \texttt{ICRL} learns such capabilities directly through in-context examples during RL rollouts—demonstrating superior data efficiency. Beyond web QA, we also evaluate \texttt{ICRL} on math reasoning tasks involving code execution as a tool. On the AIME2024 and AIME2025 benchmarks, \texttt{ICRL} matches or exceeds the performance of ReTool~\citep{feng2025retool}, a strong SFT+RL baseline, despite using no supervised pretraining. These results highlight \texttt{ICRL}'s ability to generalize to diverse tool-augmented reasoning domains and its potential as a unified, scalable framework for training tool-using models without costly supervision.

\section{In-Context Reinforcement Learning (\texttt{ICRL})}

In this section, we formally introduce tool use in LLMs and describe how RL can be applied to train such behavior. We also present the overall workflow of \texttt{ICRL}, detailing its training templates, learning process, and reward design.

\subsection{Tool Use in LLMs}

When LLMs encounter queries that exceed the scope of their internal knowledge, they must leverage external tools to obtain updated information or perform more complex reasoning. For example, search engines can provide access to recent knowledge, while Python interpreters can be used to execute structured reasoning procedures. 

Formally, given a query $q$ and an external tool $\mathcal{T}$, the model generates a response $y = (y_1, y_2, \ldots, y_{|y|})$, where each token is conditioned not only on the query and previous tokens, but also on a history of prior interactions with the tool. This defines a conditional distribution of the form:
\begin{equation}
\pi_\theta(y \mid q, \mathcal{T}) = \prod_{t=1}^{|y|} \pi_\theta(y_t \mid y_{<t}, q, \mathcal{H}_t)
\end{equation}
Here, $\pi_\theta$ is the model parameterized by $\theta$, and $\mathcal{H}_t$ denotes the sequence of previous actions taken by the model and the corresponding observations returned by the tool up to step $t$. Specifically, the interaction between the model and the tool is structured as a sequence of actions. At each time step, the model may choose to (i) perform internal reasoning, (ii) issue a query to the external tool, or (iii) return a final answer. These actions are embedded in the generated text in a structured format, such as XML tags, which distinguish reasoning steps from tool invocations and answers. For example, a reasoning step might be denoted as \texttt{<think>...</think>}, a search query as \texttt{<search>...</search>}, a retrieved information as \texttt{<information>...</information>}, and a final answer as \texttt{<answer>...</answer>}.

Furthermore, the tool functions as a response mechanism. For example, a search engine can be modeled as a retrieval function $\mathcal{T} : \mathcal{V}^* \rightarrow \mathcal{V}^*$, where $\mathcal{V}^*$ is the space of textual sequences. Given a search query $q'$, the tool returns an observation $o = \mathcal{T}(q')$, such as the top-$k$ documents retrieved from a corpus. This observation is appended to the model’s context and used in subsequent generation steps.

\subsection{RL with Tool Use} 

\paragraph{RL Objective Function.} After formulating the tool-augmented reasoning in LLMs as a Markov Decision Process (MDP), we can define the corresponding reinforcement learning (RL) objective as follows:
\begin{equation}
\begin{aligned}
\max_{\pi_\theta} \,\, 
& \mathbb{E}_{q \sim \mathcal{D},\, y \sim \pi_\theta(\cdot \mid q, \mathcal{T})} \left[ r_\phi(q, y) \right] \\
& - \beta \, \mathbb{D}_{\mathrm{KL}}\left[ \pi_\theta(y \mid q, \mathcal{T}) \,\|\, \pi_{\mathrm{ref}}(y \mid q, \mathcal{T}) \right],
\end{aligned}
\label{eq:rl}
\end{equation}
where $\pi_{\theta}$ is the policy LLM, $\pi_{\mathrm{ref}}$ is the reference LLM, $r_{\phi}$ is the reward function and $\mathbb{D}_{\mathrm{KL}}$ is KL-divergence measure. 

\paragraph{Loss Masking.}
Unlike traditional RL, which optimizes solely over model-generated tokens, tool-augmented reasoning introduces retrieved content into the rollout sequence—tokens that are not produced by the model and therefore do not reflect its internal reasoning or decision-making process. To address this, we adopt a loss masking strategy tailored for RL with tool use, which excludes retrieved content from the optimization. Specifically, only tokens generated by the language model contribute to the policy gradient, while retrieved spans are masked out and excluded from the loss computation. This targeted optimization ensures that learning remains focused on the model’s own behavior—such as tool usage, intermediate reasoning, and final answers—without being affected by fixed, untrainable content from external sources.

\paragraph{GRPO with Tool Use.} We adopt GRPO~\citep{shao2024deepseekmath} to train $\pi_{\theta}$  on the RL dataset $\mathcal{D} = \{ q_1, q_2, \cdots, q_n \}$. Specifically, for $q\in\mathcal{D}$, we use the old policy from previous step $\pi_{\theta_{\text{old}}}$ to sample a group of $N$ individual responses $\tau_i$. Then, the RL loss is defined as:
\begin{equation}
\begin{aligned}
\mathcal{L}_{\text{GRPO}}(\theta) =\; & \mathbb{E}_{\tau_i \sim \pi_{\theta_{\text{old}}}(q),\, q \sim \mathcal{D}_{\text{RL}}}
 \frac{1}{\sum_{i=1}^N |\tau_i|} \;\;\sum_{i=1}^N \sum_{t=1}^{|\tau_i|}  \\
& \text{CLIP}(r_{i,t}(\theta), A_i, \epsilon)
- \beta \cdot \mathbb{D}_{\text{KL}}[\pi_\theta \| \pi_{\text{ref}}],
\end{aligned}
\label{eq:grpo}
\end{equation}
where 
{\fontsize{8pt}{12pt}\selectfont
\begin{equation}
    A_i = \frac{R(\tau_i) - \text{mean}(\{R(\tau_i) \mid \tau_i \sim \pi_{\theta_{\text{old}}}(\tau), i = 1,2,\ldots,N\})}{\color{black}\text{std}(\{R(\tau_i) \mid \tau_i \sim \pi_{\theta_{\text{old}}}(\tau), i = 1,2,\ldots,N\})},
\end{equation}
}
and ${r_{i,t}(\theta) = {\pi_\theta(\tau_{i,t} | q, \tau_{i,<t})}/{\pi_{\theta_{\text{old}}}(\tau_{i,t} | q, \tau_{i,<t})}}$.

\subsection{\texttt{ICRL}}

\paragraph{Training Process.}

Rather than training models from scratch using reinforcement learning, which often suffers from sparse rewards and inefficient exploration, or relying exclusively on few-shot prompting, which incurs substantial inference overhead, we introduce \texttt{ICRL}, a framework that integrates the strengths of both approaches. \texttt{ICRL} leverages the sample efficiency and inductive bias of few-shot prompting while benefiting from the exploration capabilities of reinforcement learning.

\begin{table*}[t]
\centering
\caption{Few-shot rollout template in \texttt{ICRL}.}
\small
\begin{tabular}{|p{0.95\textwidth}|}
\hline
\textbf{Few-Shot Prompt Template} \\
\hline
Solve the following problem step by step. You must conduct reasoning inside \texttt{\textcolor{thinkcolor}{<think>}}...\texttt{\textcolor{thinkcolor}{</think>}} every time you get new information. After reasoning, if you find you lack some knowledge, you can call a search engine by \texttt{\textcolor{searchcolor}{<search>}} query \texttt{\textcolor{searchcolor}{</search>}} and it will return results between \texttt{\textcolor{infocolor}{<information>}}...\texttt{\textcolor{infocolor}{</information>}}. You can search as many times as you want. Finally, provide the answer inside \texttt{\textcolor{answercolor}{<answer>}}...\texttt{\textcolor{answercolor}{</answer>}}. \\
\\
Here are some examples: \\
\quad \textbf{Example Problem:} $q_{\text{demo}}$ \\
\quad \textbf{Example Solution:} \texttt{<think>}...\texttt{</think>} \texttt{<search>}...\texttt{</search>} \\
\quad\quad \texttt{<information>}...\texttt{</information>} \texttt{<think>}...\texttt{</think>} \texttt{<answer>} $a$ \texttt{</answer>} \\
\quad \textit{(repeated for $N$ examples} \\
\\
Now solve the following problem: \\
\textbf{Actual Problem:} \textcolor{questioncolor}{question} \\
\hline
\end{tabular}
\label{tab:prompt_template}
\end{table*}

At the beginning of training, we incorporate a small number of tool-use demonstrations into the model’s rollout template. These examples guide the model toward effective tool-augmented reasoning via in-context learning, akin to few-shot prompting. The resulting policy is denoted as:
\begin{equation}
\pi_\theta(y \mid \mathcal{P}_{N}, q, \mathcal{T}) = \prod_{t=1}^{|y|} \pi_\theta(y_t \mid \mathcal{P}_{N}, y_{<t}, q, \mathcal{H}_t),
\end{equation}
where $\mathcal{P}_{N}$ represents the few-shot prompt consisting of $N$ demonstration examples. Table \ref{tab:prompt_template} shows a concrete example of rollout template. 

After training for several steps, the model begins to acquire tool-use capabilities with the guidance of the initial few-shot prompt $\mathcal{P}_{N}$. Once sufficient learning progress is observed, we pause training and reduce the number of demonstration examples in the prompt. The updated policy conditioned on a reduced prompt $\mathcal{P}_{N-1}$ is defined as:
\begin{equation}
\pi_\theta(y \mid \mathcal{P}_{N-1}, q, \mathcal{T}) = \prod_{t=1}^{|y|} \pi_\theta(y_t \mid \mathcal{P}_{N-1}, y_{<t}, q, \mathcal{H}_t),
\end{equation}
where $\mathcal{P}_{N-1}$ denotes a prompt with $N-1$ demonstration examples. This process is repeated iteratively, progressively reducing the number of demonstrations, until no examples remain in the prompt.

\begin{table}[t]
\centering
\caption{Format violation penalties for computing $\mathrm{reward}_{\text{format}}$.}
\footnotesize
\setlength{\tabcolsep}{6.3pt}
\scalebox{0.95}{%
\begin{NiceTabular}{lcc}
\toprule
\textbf{Violation} & \textbf{Rationale} \\
\midrule
No \texttt{<answer>} tag  & Must provide structured answer \\
Unbalanced \texttt{<answer>} tags & Proper XML structure required \\
No \texttt{<think>} tag  & Should demonstrate reasoning \\
Unbalanced \texttt{<think>} tags & Proper XML structure required \\
No \texttt{<search>} usage  & Should utilize available tool \\
Empty answer content & Answer must be substantive \\
\bottomrule
\end{NiceTabular}%
}
\label{tab:format_penalties}
\end{table}

\paragraph{Reward Design.}

We design a composite reward function that combines the answer accuracy and format correctness to provide a richer learning signal:
\begin{equation}
    r_{\phi}(q, y) = \alpha \cdot \mathrm{reward}_{\text{acc}} + (1-\alpha)\cdot \mathrm{reward}_{\text{format}},
    \label{eq:reward}
\end{equation}
where $\alpha$ is the hyperparameter to balance two rewards. Specifically, the accuracy-based reward is computed using exact match (EM) between the model's predicted answer and the ground truth. The reward is assigned as $\mathrm{reward}_{\text{acc}} = 1$ if the prediction exactly matches the correct answer, and $0$ otherwise.

\begin{figure}[t]
\centering
\scalebox{0.9}{
\begin{minipage}{\linewidth}
\begin{algorithm}[H]
\caption{\texttt{ICRL}}
\renewcommand{\algorithmicrequire}{\textbf{Input:}}
\renewcommand{\algorithmicensure}{\textbf{Output:}}
\begin{algorithmic}[1]
  \REQUIRE Initial policy $\pi_\theta$, reference model $\pi_{\text{ref}}$, tool $\mathcal{T}$, initial few-shot prompt $\mathcal{P}_N$, dataset partitions $\{\mathcal{D}^{(N)}, \mathcal{D}^{(N-1)}, \ldots, \mathcal{D}^{(0)}\}$, reward function $r_\phi(\cdot)$, number of RL steps $T$
  \ENSURE Trained model $\pi_\theta$

  \FOR{$k = N$ to $0$}
    \STATE \texttt{// Step 1: Construct prompt with $k$ demonstrations}
    \STATE $\mathcal{P}_k \leftarrow$ select $k$ examples from $\mathcal{P}_N$
    \STATE $\mathcal{D}^{(k)} \leftarrow$ RL training subset for current prompt level

    \FOR{$t = 1$ to $T$}
      \FOR{$q \in \mathcal{D}^{(k)}$}
        \STATE $\pi_{\theta_{\text{old}}} \leftarrow \pi_\theta$
        \STATE Sample $N$ trajectories $\{\tau_1, \dots, \tau_N\} \sim \pi_{\theta_{\text{old}}}(q, \mathcal{P}_k, \mathcal{T})$

        \FOR{each trajectory $\tau_i$}
          \STATE Compute reward: $r_{\phi}(q,\tau_{i})$
          \STATE Compute normalized advantage $A_i$
          \STATE Compute importance weights $r_{i,t}(\theta)$
        \ENDFOR

        \STATE Update policy: $\pi_\theta \leftarrow \pi_\theta - \nabla_\theta \mathcal{L}_{\text{GRPO}}$
      \ENDFOR
    \ENDFOR
  \ENDFOR
\end{algorithmic}
\label{alg:icrl}
\end{algorithm}
\end{minipage}
}
\end{figure}

The $\mathrm{reward}_{\text{format}}$ component evaluates the model’s adherence to the expected structured output format, specifically the correct use of XML tags. It is defined as:
\begin{equation}
\mathrm{reward}_{\text{format}} = 1.0 - \sum_{v \in \mathcal{V}} \text{penalty}(v),
\end{equation}
where $\mathcal{V}$ denotes the set of format violations identified in the model’s response. The penalty function $\text{penalty}(v)$ assigns a predefined cost to each violation, as specified in Table~\ref{tab:format_penalties}.

With the proposed reward design, we optimize the policy using the RL objective defined in Equation~\ref{eq:grpo}. The complete training procedure for \texttt{ICRL} is outlined in Algorithm~\ref{alg:icrl}.

\begin{table*}[t]
  \centering
  \caption{Main Results of \texttt{ICRL}: Exact Match (EM) Accuracy (\%) on various difficult QA datasets. The best performance is set \textbf{bold}. The second best performance is \underline{underlined}.}
  \footnotesize
\setlength{\tabcolsep}{6.3pt}
\scalebox{0.99}{%
    \begin{NiceTabular}{c|l|ccccc|l}
      \toprule
      \multirow{2}{*}{\textbf{\normalsize{Model}}} & \multirow{2}{*}{\textbf{\normalsize{Method}}} & \multicolumn{5}{c}{\textbf{\normalsize{Difficult Question Answering}}} \vline & \multirow{2}{*}{\textbf{\normalsize{Average}}}\\
      & & TriviaQA & HotpotQA & 2Wiki & Musique & Bamboogle &  \\
      \midrule
      \multirow{12}{*}{\textbf{\normalsize{Qwen2.5-3B}}} & Direct & 28.8 & 14.9 & 24.4 & 2.0 & 2.4 & 14.50 \\
      & CoT & 3.2 & 2.1 & 2.1 & 0.2 & 0.0 & 1.52 \\
      & IRCoT & 31.2 & 16.4 & 17.1 & 6.7 & 24.0 & 19.08 \\
      & Search-o1 & 47.2 & 22.1 & 21.8 & 5.4 & 32.0 & 25.70 \\
      & RAG & 54.4 & 25.5 & 22.6 & 4.7 & 8.0 & 23.04 \\
      & SFT & 29.2 & 18.6 & 24.8 & 4.4 & 11.2 & 17.64 \\
      & R1-instruct & 44.9 & 20.8 & 27.5 & 6.0 & 19.2 & 23.68 \\
      & Reject Sampling & 48.8 & 24.0 & 23.3 & 5.9 & 21.0 & 24.60 \\
      & Search-R1 & 54.5 & \underline{32.4} & \underline{31.9} & \underline{10.3} & \underline{26.4} & \underline{31.10} \\
      & ZeroSearch & \underline{57.4} & 27.4 & 30.0 & 9.8 & 11.1 & 27.14 \\
      & \rowcolor{lightblue} \textbf{\texttt{ICRL}} & \textbf{72.6} & \textbf{35.4} & \textbf{39.2} & \textbf{20.0} & \textbf{33.6} & \textbf{40.16}${\color{+}+8.94}$ \\
      \midrule
      \multirow{13}{*}{\textbf{\normalsize{Qwen2.5-7B}}} & Direct & 40.8 & 18.3 & 25.0 & 3.1 & 12.0 & 19.84 \\
      & CoT & 18.5 & 9.2 & 11.1 & 2.2 & 23.2 & 12.84 \\
      & IRCoT & 47.8 & 13.3 & 14.9 & 7.2 & 22.4 & 21.12 \\
      & Search-o1 & 44.3 & 18.7 & 17.6 & 5.8 & 29.6 & 23.20 \\
      & RAG & 58.5 & 29.9 & 23.5 & 5.8 & 20.8 & 27.70 \\
      & SFT & 35.4 & 21.7 & 25.9 & 6.6 & 11.2 & 20.16 \\
      & R1-base & 53.9 & 24.2 & 27.3 & 8.3 & 29.6 & 28.66 \\
      & R1-instruct & 53.7 & 23.7 & 29.2 & 7.2 & 29.3 & 28.62 \\
      & Reject Sampling & 59.2 & 33.1 & 29.6 & 12.3 & 35.5 & 33.94 \\
      & Search-R1 & 61.0 & 37.0 & 41.4 & 14.6 & 36.8 & 38.16 \\
      & ZeroSearch & \underline{65.2} & 34.6 & 35.2 & 18.4 & 27.8 & 36.24 \\
      & ParallelSearch & 62.8 & \textbf{42.9} & \underline{42.4} & \underline{19.7} & \underline{41.1} & \underline{41.78} \\
      & \rowcolor{lightblue} \textbf{\texttt{ICRL}} & \textbf{75.4} & \underline{42.6} & \textbf{53.6} & \textbf{26.0} & \textbf{48.0} & \textbf{49.12}${\color{+}+7.34}$ \\
      \bottomrule
    \end{NiceTabular}%
}
\label{table:main}
\end{table*}

\section{Experiment}

\subsection{Setup}

\paragraph{Backbone Models.}

We apply \texttt{ICRL} to the Qwen2.5 model family~\citep{qwen2025qwen25technicalreport}, focusing primarily on Qwen2.5-3B-Instruct and Qwen2.5-7B-Instruct, and further evaluating on Qwen2.5-14B-Instruct. We also extend our experiments to the Qwen3 series~\citep{yang2025qwen3}, particularly Qwen3-8B, which incorporates RL enhancements. All these instruction-tuned models are widely adopted for question answering and reasoning tasks. We choose the instruct variants over base models due to their strong instruction-following capabilities, which enable faster and more stable convergence during RL training. For improved training efficiency, all models are loaded using \texttt{bfloat16} precision.

\paragraph{Baselines.} 
We evaluate the effectiveness of \texttt{ICRL} by comparing it against several state-of-the-art methods for training tool-augmented LLMs. These baselines fall into three main categories. \textbf{Direct prompting methods} include models that perform inference using direct inputs or prompting strategies such as Chain-of-Thought (CoT) reasoning~\citep{wei2022chain}. \textbf{Retrieval-based methods} leverage external information through techniques like Retrieval-Augmented Generation (RAG), including standard RAG~\citep{lewis2020retrieval}, Interleaving Retrieval Chain-of-Thought (IRCoT)~\citep{trivedi2023interleaving}, and Search-o1~\citep{li2025search}. \textbf{Fine-tuning-based methods} involve approaches such as SFT~\citep{chung2024scaling}, RL without search (R1)~\citep{guo2025deepseek}, and Rejection Sampling~\citep{ahn2024large}. We also include recent RL methods that integrate search capabilities, such as Search-R1~\citep{jin2025search}, ZeroSearch~\citep{sun2025zerosearch}, O$^2$-Searcher~\citep{mei20252}, and ParallelSearch~\citep{zhao2025parallelsearch}. These baselines provide a comprehensive comparison to validate the generality and advantages of our proposed \texttt{ICRL} framework.

\textbf{Training Datasets.} 
We use the Natural Questions (NQ) dataset~\citep{47761} as the primary training corpus. The dataset is loaded via FlashRAG~\citep{jin2025flashrag}, which provides preprocessed question-answer pairs with gold-standard answers. NQ contains real user queries from Google Search, each paired with Wikipedia passages that include the correct answer.
To support our proposed training method, we randomly sampled three questions from the web and used GPT-5.2\footnote{\url{https://platform.openai.com/docs/models/gpt-5.2}} to generate few-shot examples formatted according to the rollout template shown in Table~\ref{tab:prompt_template}. To simulate real-world tool-use behavior, we integrate the {Serper API}\footnote{\url{https://serper.dev/}} across all models to retrieve live results from the Google Search engine. For fairness, each query retrieves the top 3 documents required for search-based reasoning.

\paragraph{Evaluation Benchmarks.}
We evaluate \texttt{ICRL} and various baselines on several widely-used QA benchmarks, including TriviaQA~\citep{joshi2017triviaqa}, HotpotQA~\citep{yang2018hotpotqa}, 2Wiki~\citep{ho2020constructing}, Musique~\citep{trivedi2022musique}, and Bamboogle~\citep{press2023measuring}. Since our models are trained on the Natural Questions (NQ) dataset, we exclude NQ from the evaluation to avoid data leakage. These benchmarks cover diverse domains and reasoning types, providing a comprehensive assessment of model performance. Furthermore, to ensure evaluation efficiency, we randomly sample up to 500 questions from each dataset. The selected benchmarks include both in-domain general QA tasks (e.g., TriviaQA, HotpotQA) and out-of-domain multi-hop QA tasks (e.g., 2Wiki, Musique, and Bamboogle), allowing us to thoroughly test the generalization and reasoning capabilities of different methods.

\paragraph{Reward.} 
The hyperparameter $\alpha$ in Equation \ref{eq:reward} is set to 0.8. Furthermore, for the format violation penalties in Table~\ref{tab:format_penalties}, the weights are set to 0.5, 0.2, 0.15, 0.1, 0.1, and 0.2 respectively from top to bottom.

\begin{table*}[t]
  \centering
  \caption{Results: Exact Match (EM) Accuracy (\%) on various QA datasets. O$^2$-Searcher applies cold-start SFT before RL, while our method (ICRL) applies RL without SFT.}
  \footnotesize
\setlength{\tabcolsep}{6.3pt}
\scalebox{0.99}{%
    \begin{NiceTabular}{c|l|c|ccccc|l}
      \toprule
      \multirow{2}{*}{\textbf{\normalsize{Model}}} & \multirow{2}{*}{\textbf{\normalsize{Method}}} & \multirow{2}{*}{\textbf{\normalsize{SFT}}} & \multicolumn{5}{c}{\textbf{\normalsize{Difficult Question Answering}}} \vline & \multirow{2}{*}{\textbf{\normalsize{Average}}}\\
      & & & TriviaQA & HotpotQA & 2Wiki & Musique & Bamboogle &  \\
      \midrule
      \multirow{2}{*}{\textbf{\normalsize{Qwen2.5-3B}}} & O$^2$-Searcher & \ding{51} & 59.7 & \textbf{38.8} & 37.4 & 16.0 & \textbf{34.4} & 37.26 \\
      & \rowcolor{lightblue} \textbf{\texttt{ICRL}} & \ding{55} & \textbf{72.6} & 35.4 & \textbf{39.2} & \textbf{20.0} & 33.6 & \textbf{40.16} \\
      \bottomrule
    \end{NiceTabular}%
}
\label{table:sft}
\end{table*}

\begin{table*}[t]
\centering
\caption{Example of multi-turn search-augmented question answering. The model correctly identifies George Washington as the president who set the two-term precedent and retrieves his inauguration date through two search queries.}
\small
\begin{tabular}{|p{2.5cm}|p{12cm}|}
\hline
\textbf{Question} & When did the president who set the precedent of a two term limit enter office? \\
\hline
\textbf{Golden Answer} & April 30, 1789 \\
\hline
\textbf{Model Response} & 
\textcolor{blue}{\texttt{<think>}} I need to know which president set this precedent. I also need to know when he entered office. Let me start by finding out which president it is. \textcolor{blue}{\texttt{</think>}}

\textcolor{orange}{\texttt{<search>}} president who set two term limit precedent \textcolor{orange}{\texttt{</search>}}

\textcolor{teal}{\texttt{<information>}}
Doc 1(Title: ``Twenty-second Amendment to the United States Constitution'') Until the amendment's ratification, the president had not been subject to term limits, but both George Washington and Thomas Jefferson (the first and third ...
Doc 2(Title: ``Amendment 22 – Term Limits for the Presidency'') The decision for George Washington – the First President of the United States – to voluntarily step down from office after two terms, established an unofficial ...
Doc 3(Title: ``The first U.S. president set his own term limit'') Washington wrote ...
\textcolor{teal}{\texttt{</information>}}

\textcolor{blue}{\texttt{<think>}} It seems George Washington set the precedent. Now I need to find out when he entered office. \textcolor{blue}{\texttt{</think>}}

\textcolor{orange}{\texttt{<search>}} when did George Washington enter office as president \textcolor{orange}{\texttt{</search>}}

\textcolor{teal}{\texttt{<information>}}
Doc 1(Title: ``Presidency of George Washington'') Washington was inaugurated as the first President of the United States on April 30, 1789...
Doc 2(Title: ``George Washington'') Washington was unanimously elected ...
Doc 3(Title: ``President Washington's First Term (1789-1792)'') George Washington was inaugurated ...
\textcolor{teal}{\texttt{</information>}}

\textcolor{blue}{\texttt{<think>}} Based on the information, George Washington entered office on April 30, 1789. \textcolor{blue}{\texttt{</think>}}

\textcolor{purple}{\texttt{<answer>}} April 30, 1789 \textcolor{purple}{\texttt{</answer>}} \\
\hline
\end{tabular}
\label{tab:qa_example}
\end{table*}

\paragraph{Implementation Details.}
We implement our method using the Volcano Engine Reinforcement Learning (VeRL) framework~\citep{sheng2024hybridflow}. Qwen2.5-3B-Instruct, Qwen2.5-7B-Instruct, and Qwen2.5-14B-Instruct serve as the backbone models, trained with a learning rate of 1e-6. For each query, we sample 8 rollout trajectories with a temperature of 1.0 to compute the group-relative advantage. The maximum prompt length is set to 5000 tokens to accommodate few-shot demonstrations, and the maximum response length is capped at 2048 tokens, allowing up to 6 search turns per query.
To regularize the policy, we apply a KL penalty with a coefficient of 0.001. Training is conducted on 4 NVIDIA A100 GPUs (80GB each), using a batch size of 64. We adopt Fully Sharded Data Parallel (FSDP) training with gradient checkpointing to optimize memory usage. For retrieval, we use a BM25 retriever that returns the top-3 documents for each search query.

\begin{figure*}[htbp]
\centering
\begin{tikzpicture}
\begin{axis}[
    name=bar,
    ybar,
    bar width=8pt,
    width=0.48\textwidth,
    height=6cm,
    ylabel={EM Accuracy (\%)},
    symbolic x coords={TriviaQA, HotpotQA, 2Wiki, MuSiQue, Bamboogle},
    xtick=data,
    x tick label style={rotate=30, anchor=east, font=\small},
    ymin=0,
    ymax=100,
    legend style={
        at={(0.97,0.97)}, 
        anchor=north east, 
        legend columns=1, 
        font=\small,
        fill=white,
        fill opacity=0.8,
        draw=gray!50,
        /tikz/every even column/.append style={column sep=5pt}
    },
    legend image code/.code={
        \draw[#1] (0cm,-0.1cm) rectangle (0.3cm,0.15cm);
    },
    nodes near coords,
    nodes near coords align={vertical},
    every node near coord/.append style={font=\tiny},
    enlarge x limits=0.12,
    grid=major,
    grid style={dashed, gray!30},
    title={\small (a) EM Accuracy},
    tick style={draw=none},
]

\addplot[fill={rgb,255:red,88;green,110;blue,138}, draw={rgb,255:red,68;green,90;blue,118}] coordinates {
    (TriviaQA, 75.4)
    (HotpotQA, 42.6)
    (2Wiki, 53.6)
    (MuSiQue, 26.0)
    (Bamboogle, 48.0)
};

\addplot[fill={rgb,255:red,199;green,161;blue,130}, draw={rgb,255:red,179;green,141;blue,110}] coordinates {
    (TriviaQA, 20.8)
    (HotpotQA, 17.8)
    (2Wiki, 26.8)
    (MuSiQue, 9.0)
    (Bamboogle, 14.4)
};

\legend{3\textasciitilde2\textasciitilde0,3\textasciitilde2\textasciitilde1\textasciitilde0}
\end{axis}
\end{tikzpicture}
\hfill
\begin{tikzpicture}
\begin{axis}[
    name=line,
    width=0.48\textwidth,
    height=6cm,
    xlabel={Number of Turns},
    ylabel={Cumulative Finish (\%)},
    xmin=0.5,
    xmax=7.5,
    ymin=0,
    ymax=105,
    xtick={1,2,3,4,5,6,7},
    ytick={0,20,40,60,80,100},
    legend style={
        at={(0.97,0.35)}, 
        anchor=east, 
        font=\small,
        fill=white,
        fill opacity=0.8,
        draw=gray!50,
    },
    grid=major,
    grid style={dashed, gray!30},
    mark options={solid},
    title={\small (b) Finish Percent},
    tick style={draw=none},
]

\addplot[
    color={rgb,255:red,88;green,110;blue,138},
    mark=*,
    thick,
    mark size=2.5pt,
] coordinates {
    (1, 0.09)
    (2, 29.65)
    (3, 70.26)
    (4, 84.94)
    (5, 93.79)
    (6, 96.99)
    (7, 100.00)
};

\addplot[
    color={rgb,255:red,199;green,161;blue,130},
    mark=square*,
    thick,
    mark size=2.5pt,
] coordinates {
    (1, 23.20)
    (2, 59.39)
    (3, 83.39)
    (4, 92.80)
    (5, 97.46)
    (6, 98.68)
    (7, 99.91)
};

\legend{3\textasciitilde2\textasciitilde0, 3\textasciitilde2\textasciitilde1\textasciitilde0}
\end{axis}
\end{tikzpicture}
\caption{Comparison of Qwen-7B trained for three stages (3\textasciitilde2\textasciitilde0) vs four stages (3\textasciitilde2\textasciitilde1\textasciitilde0). (a) EM Accuracy across five QA datasets. (b) Cumulative finish percent vs number of search turns, aggregated across all datasets.}
\label{fig:combined_comparison}
\end{figure*}
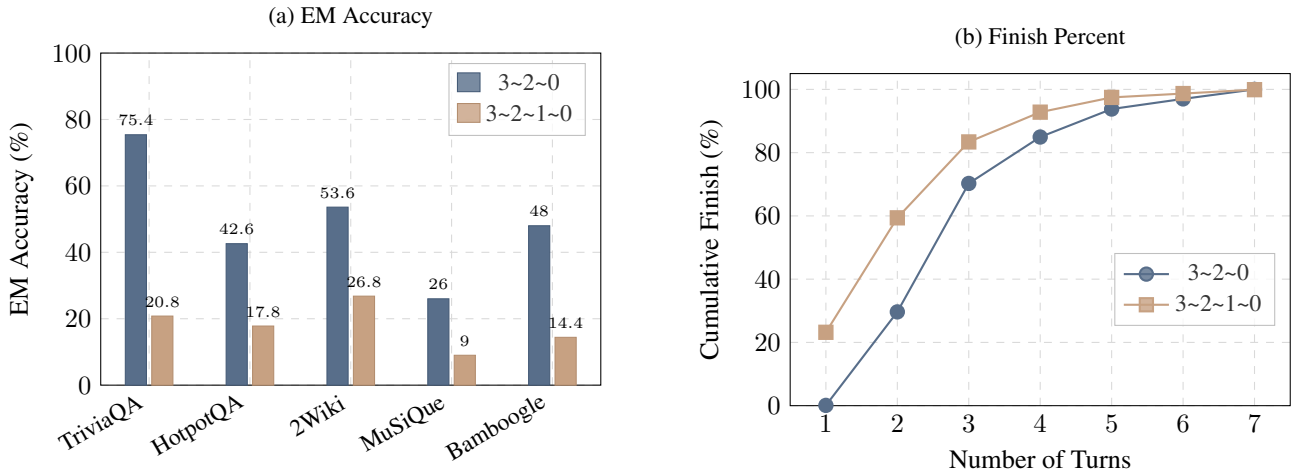

\begin{table*}[t]
  \centering
  \caption{Results: Qwen2.5-14B models Exact Match (EM) Accuracy (\%) on various QA datasets. The best performance is set \textbf{bold}.}
  \footnotesize
\setlength{\tabcolsep}{6.3pt}
\scalebox{0.99}{%
    \begin{NiceTabular}{c|l|ccccc|l}
      \toprule
      \multirow{2}{*}{\textbf{\normalsize{Model}}} & \multirow{2}{*}{\textbf{\normalsize{Method}}} & \multicolumn{5}{c}{\textbf{\normalsize{Difficult Question Answering}}} \vline & \multirow{2}{*}{\textbf{\normalsize{Average}}}\\
      & & TriviaQA & HotpotQA & 2Wiki & Musique & Bamboogle &  \\
      \midrule
      \multirow{3}{*}{\textbf{\normalsize{Qwen2.5-14B}}} & Direct & 52.0 & 22.6 & 28.2 & 6.0 & 15.2 & 24.80 \\
      & CoT & 56.4 & 24.6 & 25.8 & 9.0 & 40.0 & 31.16 \\
      & \rowcolor{lightblue} \textbf{\texttt{ICRL}} & \textbf{75.0} & \textbf{43.2} & \textbf{61.8} & \textbf{25.6} & \textbf{53.6} & \textbf{51.84} \\
      \bottomrule
    \end{NiceTabular}%
}
\label{table:14b}
\end{table*}

\subsection{Main Results}

Table \ref{table:main} presents the main results comparing \texttt{ICRL} to other baselines across five popular QA benchmarks. From the results, we can observe that: 


\paragraph{\texttt{ICRL} achieves state-of-the-art performance across QA benchmarks.}
As shown in Table~\ref{table:main}, \texttt{ICRL} significantly outperforms all baselines on both Qwen2.5-3B and Qwen2.5-7B models across five challenging QA datasets. On Qwen2.5-3B, \texttt{ICRL} achieves an average exact match (EM) score of 40.16, surpassing the best competing method, Search-R1 (31.10), by +8.94. The improvements are especially pronounced on multi-hop datasets such as 2Wiki (+7.3), Musique (+9.7), and Bamboogle (+7.2), demonstrating \texttt{ICRL}'s strength in handling complex reasoning and tool-use scenarios. 

Furthermore, on Qwen2.5-7B, \texttt{ICRL} achieves an average EM score of 49.12, outperforming the strongest baseline, ParallelSearch (41.78), by +7.34. It achieves the best results on four out of five datasets, including TriviaQA (75.4), 2Wiki (53.6), Musique (26.0), and Bamboogle (48.0). These results show that \texttt{ICRL} scales effectively with model size and generalizes well across both in-domain and out-of-domain QA tasks. The consistent gains over baselines that rely on supervised fine-tuning or reward modeling—such as ZeroSearch, Search-R1, and Reject Sampling—highlight the effectiveness of our in-context reinforcement learning framework in learning tool-use behaviors without explicit accuracy-based rewards or supervision.



\paragraph{\texttt{ICRL} achieves better performance without SFT or labeled data.}
Table~\ref{table:sft} highlights a key advantage of \texttt{ICRL}: it achieves superior performance without requiring any SFT, in contrast to O$^2$-Searcher, which applies a cold-start SFT phase before reinforcement learning. Despite using no labeled tool traces or task-specific supervision, \texttt{ICRL} achieves a higher average EM score of 40.16 compared to 37.26 from O$^2$-Searcher. It outperforms O$^2$-Searcher on four out of five datasets, including substantial gains on TriviaQA (+12.9) and Musique (+4.0). These results demonstrate that \texttt{ICRL} can learn effective tool-use strategies purely from in-context examples and reinforcement signals, offering a scalable and data-efficient alternative to methods that rely on costly annotation and pretraining.

\subsection{Concrete Examples}



Table~5 presents a complete reasoning example from \texttt{ICRL}-Qwen2.5-7B on a question from the {Bamboogle} dataset. The model is tasked with answering a compositional query: identifying the president who established the two-term precedent and determining when he entered office. It first issues a search query to identify the relevant figure, correctly concluding that George Washington set the precedent. It then formulates a follow-up query to retrieve his inauguration date and successfully extracts the correct answer, April 30, 1789.
This case illustrates \texttt{ICRL}’s ability to decompose complex questions, retrieve relevant information across multiple turns, and maintain coherent reasoning without explicit intermediate supervision. It demonstrates the effectiveness of our framework in learning structured tool-use behaviors through in-context reinforcement learning.

\section{Further Analysis}

\begin{figure*}[htbp]
    \centering
    \begin{subfigure}[b]{0.32\textwidth}
        \centering
        \includegraphics[width=\textwidth]{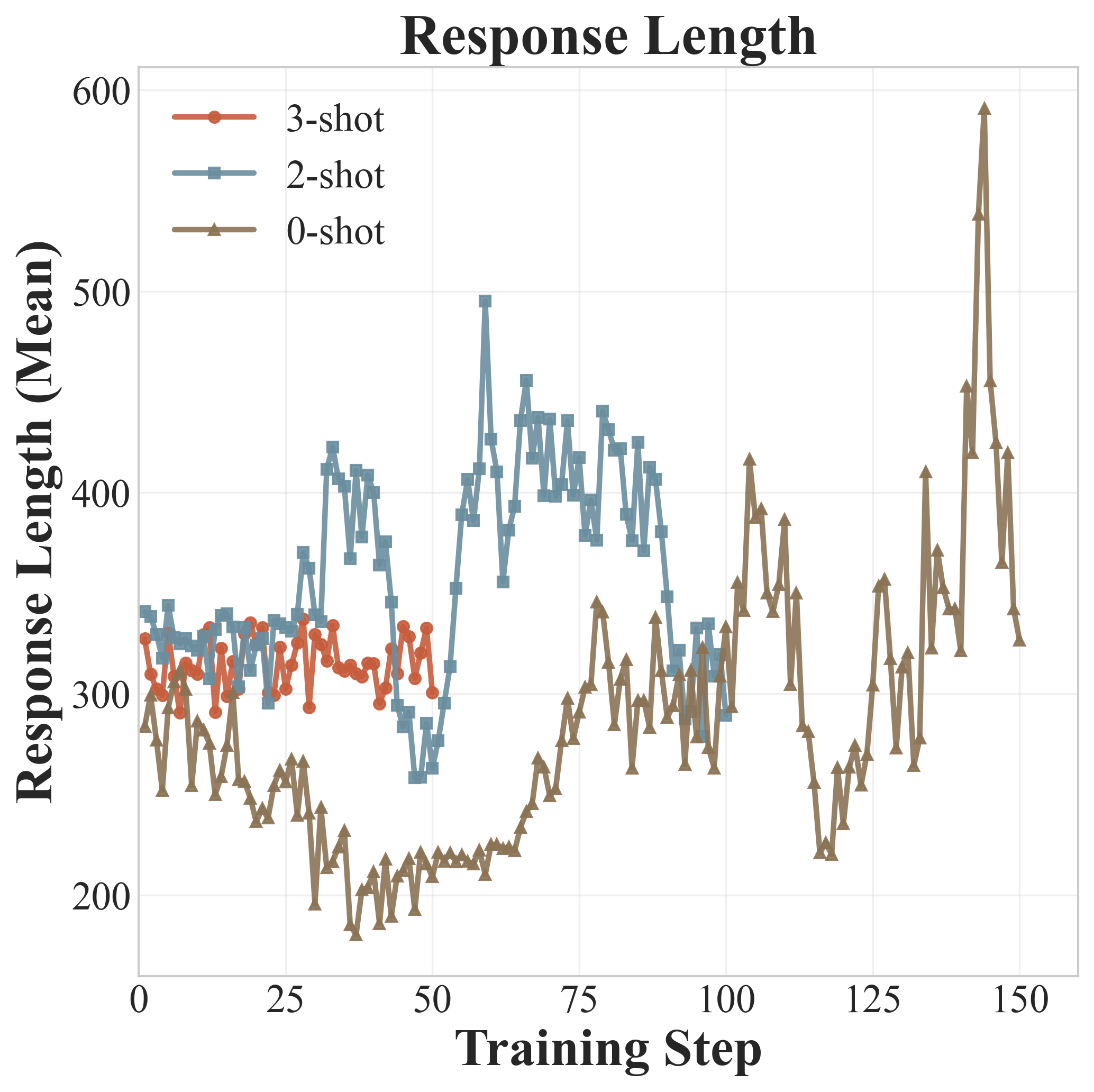}
        \caption{Response Length}
        \label{fig:response_length}
    \end{subfigure}
    \hfill
    \begin{subfigure}[b]{0.32\textwidth}
        \centering
        \includegraphics[width=\textwidth]{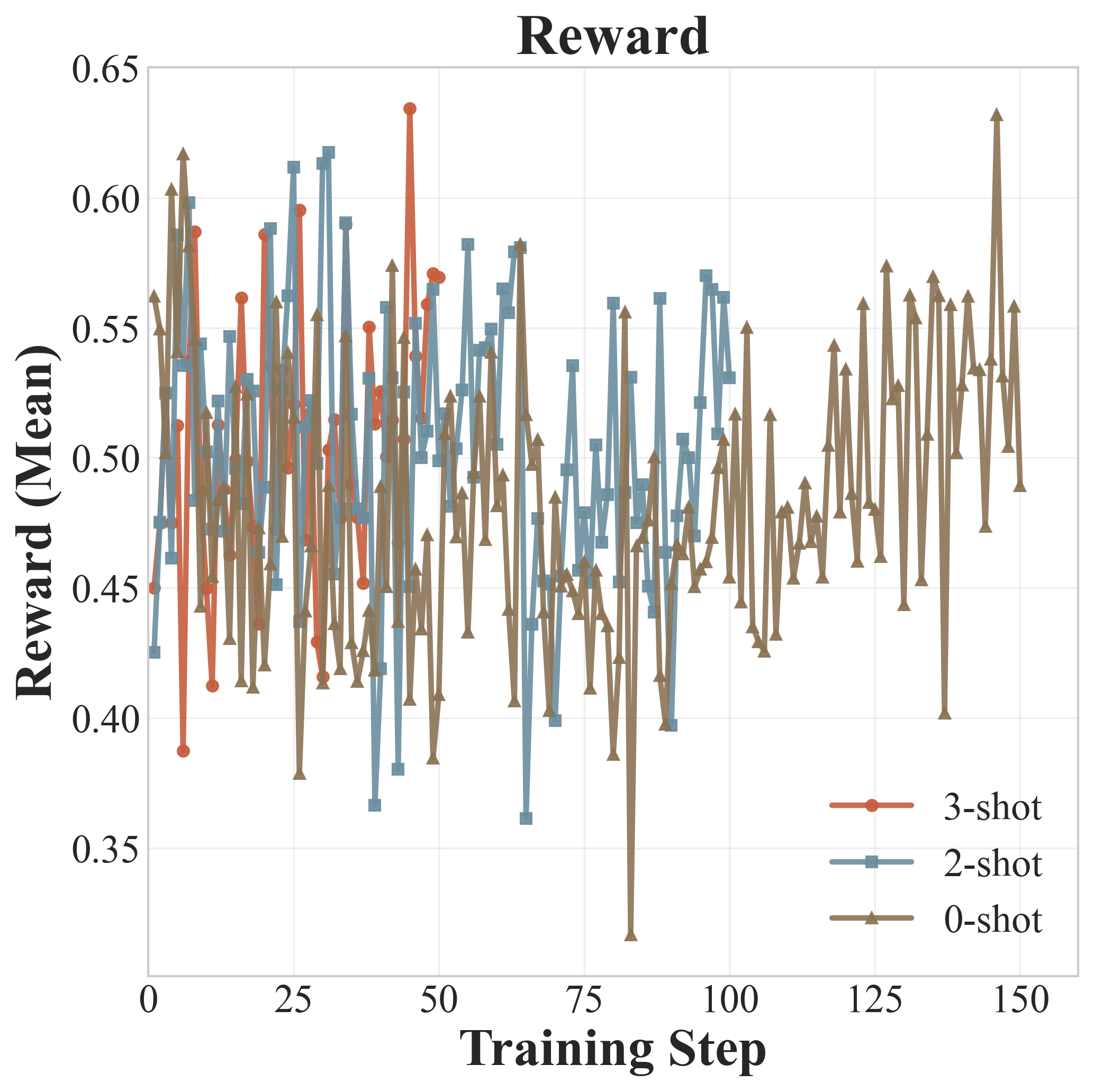}
        \caption{Reward}
        \label{fig:reward}
    \end{subfigure}
    \hfill
    \begin{subfigure}[b]{0.32\textwidth}
        \centering
        \includegraphics[width=\textwidth]{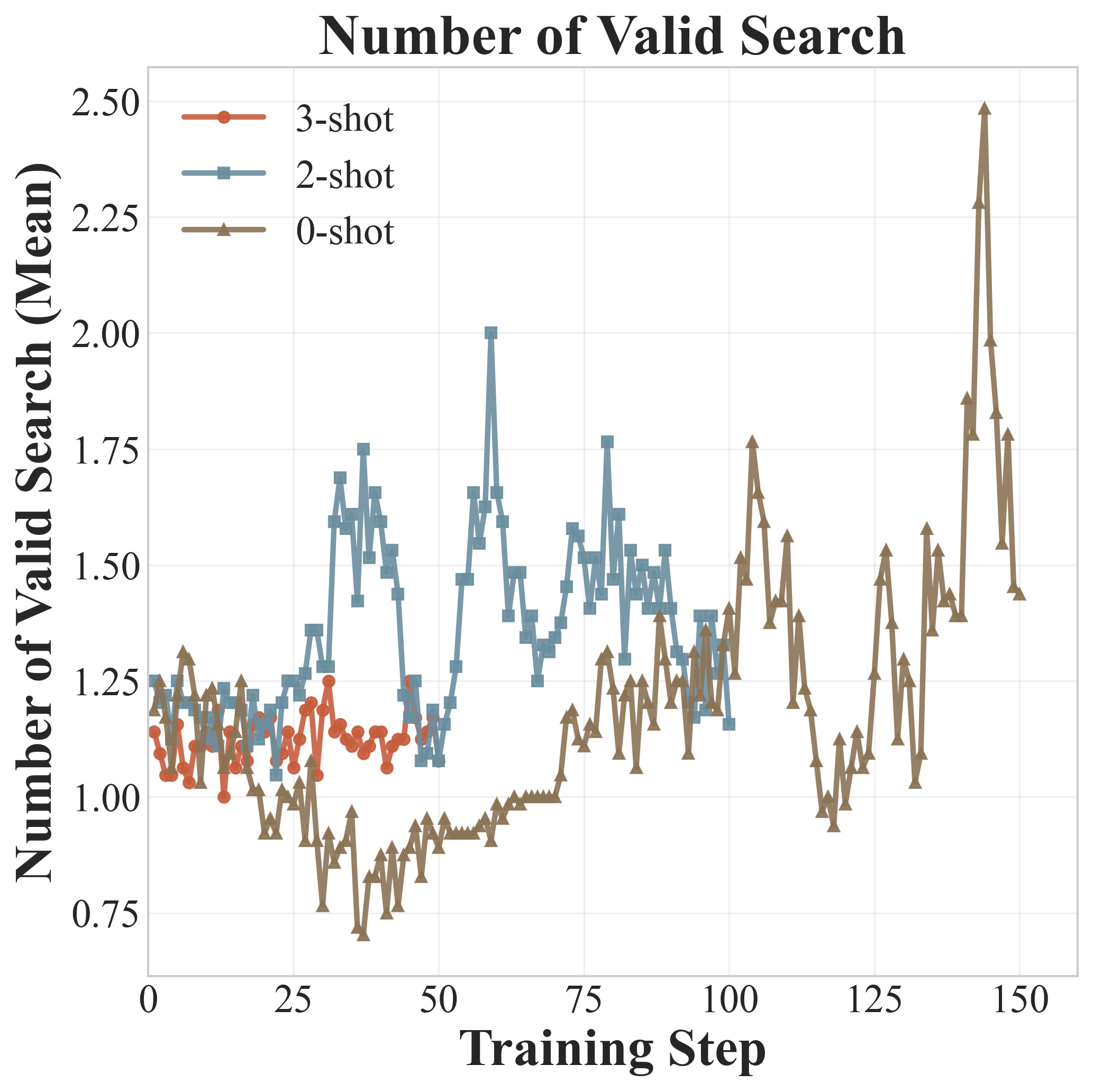}
        \caption{Number of Valid Search}
        \label{fig:valid_search}
    \end{subfigure}
    \caption{Training dynamics comparison across different few-shot settings (3-shot, 2-shot, 0-shot) for Qwen-7B model.}
    \label{fig:training_dynamics}
\end{figure*}

\subsection{Ablation analysis}

\paragraph{Ablation on curriculum design for rollout reduction.}
We conduct an ablation study comparing two curricula for reducing the number of examples used in the rollout process: a three-stage schedule (3\textasciitilde2\textasciitilde0) and a four-stage schedule (3\textasciitilde2\textasciitilde1\textasciitilde0). As shown in Figure~\ref{fig:combined_comparison} (a), the three-stage variant achieves substantially higher EM accuracy across all five QA datasets. For instance, on TriviaQA and 2Wiki, the three-stage model reaches 75.4 and 53.6, compared to 20.8 and 26.8 with the four-stage version. 

Figure~\ref{fig:combined_comparison} (b) shows that the four-stage curriculum leads to faster decisions, with over 80\% of queries finishing within two search turns. However, this comes at the cost of answer quality. These results suggest that aggressively reducing rollout length too early (via the intermediate stage with one example) encourages premature stopping and weakens multi-turn reasoning. In contrast, the simpler 3\textasciitilde2\textasciitilde0 curriculum maintains stronger performance by allowing the model to explore longer reasoning paths during training.




\paragraph{Model Scaling and 14B Performance.}
To evaluate the scalability of \texttt{ICRL} across larger models, we applied our method to Qwen2.5-14B-Instruct and report results in Table~\ref{table:14b}. \texttt{ICRL} significantly outperforms both direct prompting and CoT methods across all five QA datasets. In particular, it achieves 75.0 EM on TriviaQA and 61.8 on 2Wiki, yielding a strong average EM score of 51.84—surpassing CoT by +20.7 and direct prompting by +27.0. These results demonstrate that \texttt{ICRL} continues to scale effectively to larger model sizes and benefits from increased capacity, without requiring additional supervision or annotation.

\subsection{Training process}


To understand how \texttt{ICRL} evolves during training, we analyze the learning curves across the 3-shot, 2-shot, and 0-shot curriculum stages using the Qwen2.5-7B model. In the early stages with demonstrations (3-shot and 2-shot), the model produces relatively stable and well-structured responses, as reflected in the consistent response lengths. As training progresses into the 0-shot stage, the response length initially drops due to the removal of in-context examples but gradually increases again, indicating that the model is learning to independently compose longer and more structured outputs.

Although the reward remains relatively steady throughout training and is based only on sparse signals—output format validity and final answer accuracy—the model still learns to use tools more effectively over time. This is most clearly reflected in the increasing number of valid tool calls during the 0-shot phase. The rise in valid tool usage indicates that \texttt{ICRL} successfully encourages the model to internalize tool-use behavior, even without dense or step-level supervision.

\subsection{Generalize}

\begin{table}[t]
  \centering
  \caption{Results: Accuracy (\%) on Math QA datasets.}
  \footnotesize
  \setlength{\tabcolsep}{6.3pt}
  \begin{NiceTabular}{c|l|c|cc}
    \toprule
    \multirow{2}{*}{\textbf{\normalsize{Model}}} & \multirow{2}{*}{\textbf{\normalsize{Method}}} & \multirow{2}{*}{\textbf{\normalsize{SFT}}} & \multicolumn{2}{c}{\textbf{\normalsize{Math QA}}} \\
    & & & AIME2024 & AIME2025 \\
    \midrule
    \multirow{2}{*}{\textbf{\normalsize{Qwen3-8B}}} & ReTool & \ding{51} & 67.0 & 49.3  \\
    & \rowcolor{lightblue} \textbf{\texttt{ICRL}} & \ding{55} & 64.1 & 51.7 \\
    \bottomrule
  \end{NiceTabular}
  \label{table:math}
\end{table}

Apart from conducting our method in web search tool-calling domain, we also evaluate our method by training the models' abilities on code-writing and calling tools to run the python code to help them solve complex math problems. We compare our results with ReTool, which is a SFT-RL training framework that training models to learn code-writing and tool-calling on running code. ReTool~\citep{feng2025retool} achieves the state-of-the-art performance on code-augmented long-form reasoning for solving math problems, but it also needs a lot of annotated data to apply cold-start SFT for models to learn the tool-calling format and understand the whole reasoning process. However, our method \texttt{ICRL} doesn't need to SFT the model first, and models can still reasoning properly from the In-Context Learning with the examples we provides in prompt. From Table \ref{table:math}, althrough out method underperforms ReTool on AIME2024 by 2.9\%, it can achieve better result on AIME2025 with +2.4\% accuracy. Which means our method still works for help models to learn other tool-calling operations and is more data-efficient than other methods that need cold-start SFT with thousands of annotated data.

\section{Conclusion}

We introduce \texttt{ICRL}, a simple yet powerful framework for training LLMs to use tools via in-context reinforcement learning, without requiring SFT or labeled tool traces. By incorporating few-shot demonstrations directly into the RL rollout prompts and gradually phasing them out, \texttt{ICRL} enables models to transition from imitation to autonomous tool use through reward-driven learning.
Our method achieves strong performance across a range of QA and reasoning benchmarks, outperforming existing approaches that rely on supervised data or frozen tool-use policies. \texttt{ICRL} also generalizes across domains, including web search and code execution, demonstrating its flexibility and effectiveness. These results highlight \texttt{ICRL} as a scalable and data-efficient alternative to traditional SFT+RL pipelines for enabling tool-augmented language models.

\section*{Impact Statements}


This paper presents work whose goal is to advance the field of machine learning. There are many potential societal consequences of our work, none of which we feel must be specifically highlighted here.


\bibliography{example_paper}
\bibliographystyle{icml2026}

\newpage
\appendix
\onecolumn

\end{document}